\documentclass[letterpaper,10 pt, conference]{ieeeconf}  

\IEEEoverridecommandlockouts                              
\usepackage{cite}
\usepackage{graphicx}
\usepackage{caption}
\usepackage{booktabs}
\usepackage{float}
\usepackage{mathtools, cases}
\usepackage{amsmath,empheq}
\usepackage{eqparbox} 
\usepackage{bm} 
\usepackage{multirow}
\usepackage{lipsum} 
\usepackage{subcaption}
\usepackage{mwe}
\usepackage{tikz,lipsum}

\title{\LARGE \bf
PPGnet: Deep Network for Device Independent Heart Rate Estimation from Photoplethysmogram
}

\author{Shyam A, Vignesh Ravichandran, Preejith S.P, Jayaraj Joseph and Mohanasankar Sivaprakasam 
\thanks{Shyam A, Vignesh Ravichandran, Preejith S.P, Jayaraj Joseph and Mohanasankar Sivaprakasam are with the Healthcare Technology Innovation Centre (HTIC),       Indian Institute of Technology Madras, India }
\thanks{(e-mail:shyam@htic.iitm.ac.in)}
\thanks{Mohanasankar Sivaprakasam is with the Department of Electrical Engineering, Indian Institute of Technology Madras, India}%
}

\begin{document}

\maketitle
\thispagestyle{empty}
\pagestyle{empty}

\begin{abstract}
Photoplethysmogram (PPG) is increasingly used to provide monitoring of the cardiovascular system under ambulatory conditions. Wearable devices like smartwatches use PPG to allow long-term unobtrusive monitoring of heart rate in free-living conditions. PPG based heart rate measurement is unfortunately highly susceptible to motion artifacts, particularly when measured from the wrist. Traditional machine learning and deep learning approaches rely on tri-axial accelerometer data along with PPG to perform heart rate estimation. The conventional learning based approaches have not addressed the need for device-specific modeling due to differences in hardware design among PPG devices. In this paper, we propose a novel end-to-end deep learning model to perform heart rate estimation using 8-second length input PPG signal. We evaluate the proposed model on the IEEE SPC 2015 dataset, achieving a mean absolute error of 3.36$\pm$4.1BPM for HR estimation on 12 subjects without requiring patient-specific training. We also studied the feasibility of applying transfer learning along with sparse retraining from a comprehensive in-house PPG dataset for heart rate estimation across PPG devices with different hardware design.
\end{abstract}

\section{INTRODUCTION}

Heart rate is one of the four primary vital signs that indicate the general wellbeing of an individual. The traditional method to estimate heart rate was using Electrocardiogram (ECG) signal,  however, the challenges with its long-term usage and comfort have resulted in the adoption of Photoplethysmogram (PPG) \cite{parak2014evaluation}. PPG measures the volumetric change of blood in a given location by measuring variations in the amplitude of transmitted or reflected light under the illumination of either a visible or infrared light source. PPG can be measured from multiple sites such as finger \cite{avolio2002finger}, earlobe, forehead and more commonly in wearables, the wrist. PPG is inexpensive, compact and comfortable for long-term use and through its design provides galvanic isolation, making it an ideal choice for wearable devices. By measuring the changes in the PPG signal, heart rate can be derived. However, heart rate computation in PPG is highly susceptible to motion artifacts. Motion artifacts in particular affects the utility of heart rate measurement during physical activities like exercise.

\quad \quad Traditional filtering techniques provide poor performance in filtering out motion artifacts as they lack a specific frequency range and often overlap with the frequency band of the PPG waveform \cite{shimazaki2014cancellation}. 
Recent advances in the topic of motion artifact reduction in PPG has largely been driven by the IEEE Signal Processing Cup (SPC) 2015 database accompanying the TROIKA framework which provided PPG, ECG and accelerometer recordings of individuals under physical load. 
Feature-based methods and adaptive filtering methods have been proposed for heart rate estimation using the IEEE SPC 2015 database \cite{chowdhury2018real}. These methods require apriori knowledge about the motion artifact profiles and require significant effort to function with alternate datasets.

\quad\quad  Deep learning based methods have also been proposed for the same. Biswas et al. \cite{biswas2019cornet} propose a combination of Convolution Neural Networks (CNN) and Long Short Term Memory (LSTM) to perform both heart rate estimation and biometric identification using raw input PPG. The approach, however, requires patient-specific retraining which results in additional hurdles during deployment. Reiss et al. \cite{reiss2018ppg} applied Fourier Transform on the input PPG and tri-axial accelerometer data before passing it through a CNN based deep learning model. This introduces an intensive preprocessing step and results in more number of parameters. Both the learning based and conventional methods make use of the tri-axial accelerometer data along with PPG signal for heart rate estimation. This, however, imposes a significant battery load on wearable devices like smartwatches, due to the requirement of either processing or transmitting the additional accelerometer data. None of the learning based approaches proposed earlier have shown their ability to generalize across different datasets requiring exclusive training with the new dataset, which greatly hampers the utility of such a solution. The differences in hardware specifications between wearable PPG devices causes variances among the PPG signals while estimating heart rate. A similar problem was addressed in the domain of image classification using deep learning networks where transfer learning allowed for the reuse of feature maps across different datasets causing a reduction in dataset size and training time\cite{shin2016deep}. A similar approach is required to address the variability between different PPG devices. 

\quad\quad In this paper, we propose a deep learning model for heart rate estimation using a single channel wrist PPG signal that was not dependent on patient-specific training. We also studied the feasibility of performing transfer learning using model trained on the IEEE SPC 2015 to perform heart rate estimation through sparse sampling from a large in-house dataset collected using custom hardware.

In summary, the contributions in this paper are as follows:

\begin{itemize}[noitemsep,partopsep=0pt]
\item {We propose a novel end-to-end deep learning architecture to perform estimation of heart rate using single channel wrist PPG.}
\item{We conduct an extensive evaluation of the proposed model on both the IEEE SPC 2015 dataset and the in-house dataset.}
\item{We study the feasibility of using transfer learning with a model trained on the IEEE SPC 2015 dataset and retrained on a sparse subset of the in-house dataset to pave the way for the development of device agnostic deep learning models for heart rate estimation.}
\end{itemize}

\section{DATASET DESCRIPTION}
\begin{figure*}[!htb]
\centering
\includegraphics[width=\textwidth]{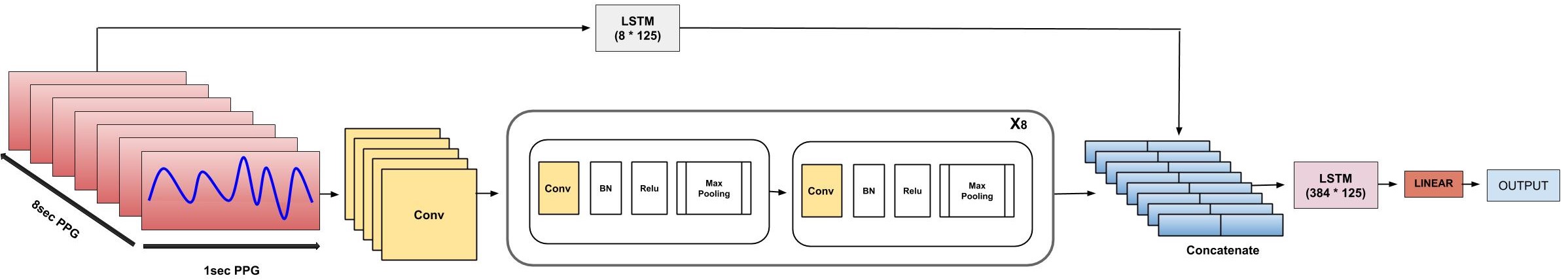}
\caption{Proposed PPGnet Architecture with CNN and LSTM Components}
\label{Architecture}
\end{figure*} 

\subsection{IEEE SPC 2015 dataset}
The IEEE SPC 2015 dataset consists of two separate groups, one collected from 12 subjects (IEEE Training), who were made to run on a treadmill at varying speed and the second subset collected from 8 subjects (IEEE Test), who were made to perform exhaustive arm movements like forearm exercise and boxing. The data was collected using a wrist-worn device to acquire two-channel PPG, tri-axial acceleration using an accelerometer and a single lead ECG. All the signals were sampled at 125 Hz and recorded simultaneously. Additional details regarding the data collection protocol can be found in \cite{zhang2015troika}. The wide range of physical motion involved through the course of the study was crucial to analyze the robustness of the proposed model. This study only used the first group for training and evaluation of the proposed model. In the later sections, the first group would be referred to as the IEEE SPC 2015 dataset.

\subsection{ADI dataset}

A comprehensive in-house data collection of PPG was conducted under two different conditions. The first test condition was aimed to inspect the changes in PPG signal during induced stress through Trier Social Stress Test (TSST) procedure \cite{kirschbaum1993trier}. The second test condition was aimed to capture PPG data under randomized motion where each subject was allowed to perform their daily activities like walking, sitting, computer usage, hand gestures, and other random movements to provide more variability in the data. Further details about the study protocol can be seen from our previous work. The dataset would hereafter be referred to as the ADI dataset. \cite{amalan2018electrodermal}. 
These conditions capture a wide range of heart rates that encompasses events ranging from physical stress, mental stress and daily activity. 
The data was collected from 50 subjects (36 male 14 female) using a wrist-worn device from Analog Devices, Inc. PPG (Green LED), an accelerometer at 50 Hz and ECG at 400 Hz were recorded. 
\begin{table}[!htb]
\centering
\caption{Dataset description }
%
\begin{tabular}{|c|c|c|}
\hline
Dataset & \begin{tabular}[c]{@{}c@{}}Number of subjects\end{tabular} & \begin{tabular}[c]{@{}l@{}}Total number of windows\end{tabular} \\ \hline
ADI & 50 & 22808 \\ \hline
IEEE SPC training & 12 & 1765 \\ \hline
IEEE SPC test & 10 & 1327 \\ \hline
\end{tabular}
\label{windows_dataset}
\end{table}

\subsection{Dataset Preparation}
The ADI dataset for training the proposed model was prepared in accordance with the IEEE SPC 2015 dataset where the heart rate labels were provided for every 8-second window of the PPG. Further,  the PPG signal was upsampled to 125 Hz. Each subject data was segmented into 8-second windows by sliding window approach with an overlap of 6 seconds (shift by 2 seconds). The total number of windows formed from each dataset is listed in Table \ref{windows_dataset}. Ground truth was obtained by calculating the mean heart rate for the corresponding ECG windows. A second-order band-pass Butterworth filter with cut-off frequency of 0.5-5 Hz was applied to the raw PPG windows to remove baseline wander and high frequency artifacts. PPG windows were normalized on subject basis.     

\section{METHODOLOGY}
\subsection{Network design}
The proposed deep learning network is shown in Fig. \ref{Architecture} contains three components namely the Convolution Neural Networks (CNN), Long Short Term Memory (LSTM) and Fully Connected Network (FCN).
The input data is segmented into 8 1-second windows and passed to both the CNN and LSTM feature extractors. To the CNN feature extractor, each PPG window of length 125 samples were fed into the inception block which performs 5 parallel convolution using kernels of size (1x5, 1x20, 1x40, 1x60, 1x80). These provide diverse feature representations from the input signal at various receptive fields. The output features from the inception block are concatenated together and passed into two sequential blocks, each containing convolution block followed by Batch Normalization (BN), ReLU, Max-Pooling and Dropout. 

\quad \quad The input data is also provided to an LSTM for ensembling with the features extracted by CNN \cite{murugesan2018ecgnet}. The LSTM feature extractor provides features with temporal dependencies from the given input PPG. The nature of the heart rate estimation task greatly benefits from incorporating time-dependent features. 
The CNN and LSTM features corresponding to each 1-second input window are concatenated and provided as an input to another LSTM at each of the 8 time-steps. The final time step hidden features of the second LSTM is used as the input to a fully connected layer which outputs a scalar value of the estimated heart rate. The key parameters involved in the design of the proposed model is shown in Table \ref{model_desc}.

\begin{table}[]
\centering
\caption{Description of proposed model parameters}
\resizebox{0.445\textwidth}{!}{%
\begin{tabular}{|c|c|c|c|c|}
\hline
\textbf{Block} & \textbf{Component} & \textbf{Kernel size} & \textbf{Input} & \textbf{Output} \\ \hline
\multirow{2}{*}{Sequential block 1} & Convolution & 40 & 16 & 32 \\ \cline{2-5} 
 & Max-pooling & 4 & - & - \\ \hline
\multirow{2}{*}{Sequential block 2} & Convolution & 60 & 32 & 32 \\ \cline{2-5} 
 & Max-pooling & 4 & - & - \\ \hline
 \textbf{Block} & \textbf{Component} & \textbf{Input} & \textbf{Hidden} & \textbf{Layers} \\ \hline
\multirow{2}{*}{LSTM} & LSTM 1 & 125 & 80 & 2 \\ \cline{2-5} 
 & LSTM 2 & 384 & 80 & 2 \\ \hline
\end{tabular}
\label{model_desc}

}
\end{table}


\subsection{Implementation}
All the models were implemented in PyTorch, version 0.4.0. The model training was carried out on an Nvidia GTX 1080 Ti GPU with 12 GB RAM for 750 epochs using a batch size of 128. The network was initialized with random weights and the learning rate was fixed as 0.02. Stochastic Gradient Descent (SGD) was used as the optimizer and loss was computed as the absolute difference between ground truth and estimated heart rate values.

\subsection{Validation of datasets}
There have been different approaches in validating the PPG signal for heart rate estimation. As reported by Reiss et al. \cite{reiss2018ppg}, a significant performance difference is expected between subject or session dependent and session independent evaluation. Our focus in this paper was to show the model{'}s capability to generalize across different datasets. We find that cross-validation using both Leave One Session Out (LOSO) and five-fold cross-validation scheme were apt for validating IEEE SPC 2015 and ADI datasets independently. The performance metric used on any given dataset was estimated by computing Mean Absolute Error (MAE) between actual and predicted heart rates. 
\begin{equation}
MAE = \sum_{i=1}^{m}|BPM_{actual}(m) - BPM_{predicted}(m)|
\end{equation}	
where $BPM$ is beats per minute and $m$ is the total number of windows.
\subsection{Feasibility of applying transfer learning}
In addition to prior works, we studied the feasibility of utilizing transfer learning in the proposed model trained on the IEEE SPC 2015 dataset followed by sparse retraining on the ADI dataset. Transfer learning has traditionally been used in image classification tasks where it was reported that when training a model on a new dataset by optimizing the penultimate feature layers alone provided similar performance and parameter reduction. This is due to the fact that early feature representations require only minor modification to function in different datasets as shown in image classification networks. Similarly, this idea can be used for the PPG based heart rate estimation task, due to the fundamental similitude of the PPG signal morphology across different datasets. 

\quad \quad The variance between different datasets can be attributed to a wide range of factors from:
\begin{itemize}
\item{Measurement site}
\item{Mode of acquisition (Reflectance, Transmittance)}
\item{Light source wavelength}
\item{Source-Detector distance}
\item{ADC resolution \& gain}
\end{itemize} 

\quad \quad To account for the aforementioned sources of variance among datasets, sparse retraining of the penultimate features is required to permit the model to function in new datasets. Evaluation on the ADI dataset was carried out in four distinct conditions. In the first condition, five-fold cross-validation was conducted on the model exclusively trained and evaluated on the ADI dataset. In the second condition, the model was trained on the IEEE SPC 2015 dataset and evaluated on the ADI dataset. 
For the third condition, five-fold cross-validation was used on the ADI dataset with pretrained model weights from IEEE SPC 2015 dataset while optimizing the $LSTM2$ and $Linear$ blocks. In the fourth condition, retraining was carried out similar to the third condition, but only through the use of a sparse training data (15\%) of the ADI dataset and tested on the remaining data.  

\begin{figure}%
\centering
\includegraphics[width=0.445\textwidth]{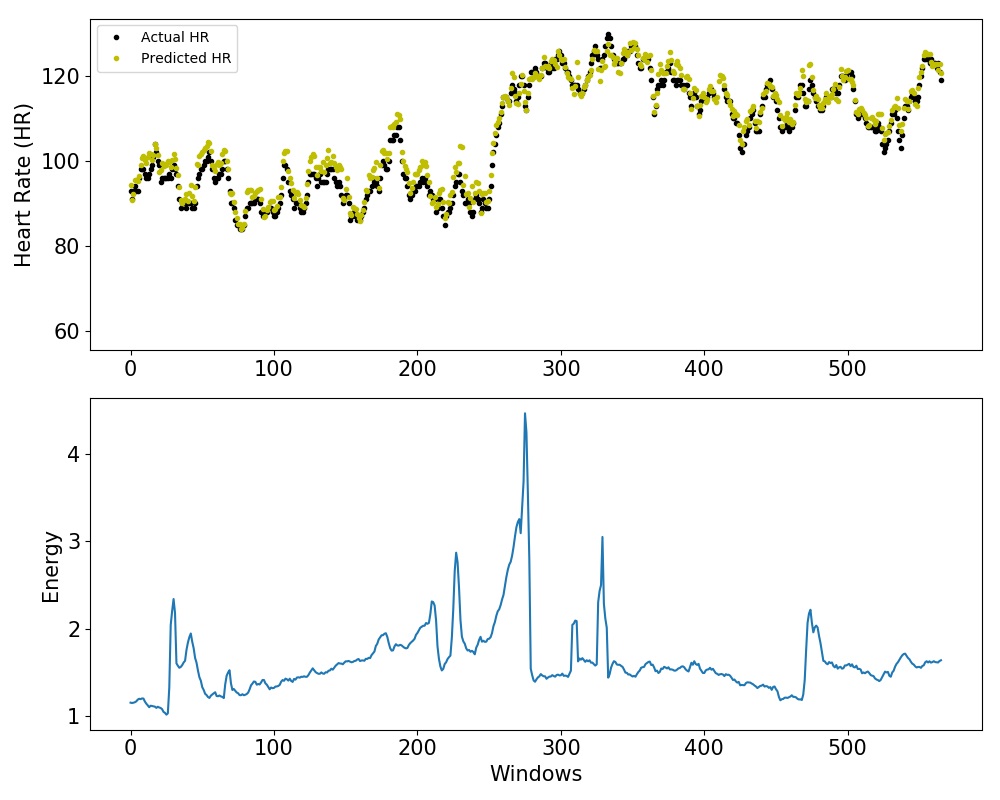}%
\caption{Evaluation of an example ADI dataset}
\label{subfig4a}%
\setlength{\belowcaptionskip}{-10pt}
\end{figure}

\section{ANALYSIS AND RESULTS}
\subsection{IEEE SPC 2015}
The model was first validated against IEEE SPC 2015 dataset. The Mean Absolute Error (MAE) and the Standard Deviation of Absolute Error (SDAE) were observed to be 3.36$\pm$4.1 (MAE$\pm$SDAE), which is comparable to state-of-the-art methodologies, Table \ref{stateofart}. 
The correlation coefficient for all the subjects was found to be 0.98. 
The IEEE SPC test dataset was also validated to check the model performance under intense physical movements and it had a mean error of 12.48$\pm$14.45 (MAE$\pm$SDAE).

\begin{table}[]
\centering
\caption{Result comparison with state-of-the-art methods}
\label{stateofart}
\resizebox{0.5\textwidth}{!}{%
\begin{tabular}{|c|c|c|c|c|}
\hline
\textbf{Network} & \textbf{IEEE Training} & \textbf{IEEE Testing} & \textbf{Signals used} & \textbf{Preprocessing} \\ \hline
SpaMa\cite{salehizadeh2015novel} & 13.1$\pm$20.7 & 9.20$\pm$11.4 & PPG and accelerometer & Yes \\ \hline
SpaMaPlus\cite{reiss2018ppg} & 4.25$\pm$5.9 & 12.31$\pm$15.5 & PPG and accelerometer & Yes \\ \hline
Schaeck2017\cite{schack2017computationally} & 2.91$\pm$4.6 & 24.65$\pm$24 & 2 channel PPG & Yes \\ \hline
Reiss et al.\cite{reiss2018ppg} & 3.91 & 18.33 & PPG and accelerometer & Yes \\ \hline
\textbf{PPGnet} & 3.36$\pm$4.1 & 12.48$\pm$14.45 & Single channel PPG & No \\ \hline
\end{tabular}
}
\end{table}

\begin{table}[]
\centering
\caption{Evaluation of network performance under different conditions on the ADI dataset}
\label{ADI}
\resizebox{0.5\textwidth}{!}{%
\begin{tabular}{|c|c|c|c|c|c|}
\hline
\textbf{Condition} & \textbf{Initial Weights} & \textbf{Parameters} & \textbf{Epochs} & \textbf{MAE$\pm$SDAE} & \textbf{PCC} \\ \hline
Condition 1 & Random & 765265 & 750 & 4.10$\pm$5.19 & 0.92 \\ \hline
Condition 2 & Random & 765265 & - & 18.77$\pm$20.21 & 0.41 \\ \hline
Condition 3 & IEEE SPC 2015 & 391841 & 65 & 4.095$\pm$5.25 & 0.92 \\ \hline
Condition 4 & IEEE SPC 2015 & 391841 & 90 & 4.11$\pm$5.27 & 0.92 \\ \hline
\end{tabular}
}
\end{table}
\subsection{ADI Dataset}
\vspace{-1pt}
The ADI dataset covers a wide range of heart rates ranging from 39 to 155 BPM which is far more constitutive of heart rates encountered in real life conditions. Table \ref{ADI} shows the network performance under different testing conditions. The network achieved the highest performance when exclusively trained and tested on the ADI dataset, but it requires over 750 epochs to converge. When the network was trained on IEEE SPC 2015 dataset and tested on ADI, poor results were observed due to the variance between the PPG device design in the datasets that was not accounted for during evaluation. For the third condition, the network was trained on ADI dataset with pretrained model weights of IEEE SPC 2015 dataset. It was observed that convergence is attained within 65 epochs to attain a similar performance to the model exclusively trained on the ADI dataset. Finally, it can be observed that even when training using a subset of ADI dataset initialized with pretrained model weights of IEEE SPC 2015 dataset, the convergence was achieved in 90 epochs with similar performance of the first test condition. This indicates the feasibility of developing device-specific models through minimal retraining as opposed to exclusive training on different datasets. 

\quad \quad Further analysis was done on the model trained with the minimal ADI dataset. The model achieved an average error of 4.10$\pm$5.25 (MAE$\pm$SDAE) with a Pearson Correlation Coefficient (PCC) of 0.92.
Randomized motion poses a significant challenge for estimating heart rate. Fig. \ref{subfig4a} shows the performance of the proposed model in estimating heart rate under randomized motion along with the average root mean square value of the accelerometer axis. It can be observed that, despite the high energy found in the accelerometer signal between 250 to 400 windows, the model was able to adapt to the quick transition, making it less prone to motion artifacts. 

\section{CONCLUSIONS}

In this paper, we propose a novel end-to-end deep learning model to perform the task of heart rate estimation using ambulatory wrist PPG signal which does not require extensive preprocessing or patient-specific training. We report an average error of 3.36$\pm$4.1 (MAE$\pm$SDAE) on the IEEE SPC 2015 dataset.
We studied the feasibility of using transfer learning by reporting similar results between sparse retraining and exclusive training on the ADI dataset.
This suggests that the proposed model can adapt to PPG devices with different hardware specifications requiring only sparse training for a new hardware device. Extensive evaluation is required to test the ability of the proposed model to handle different sources of variance in PPG data like light source wavelength, multiple sources, different sites and skin tone. 

\addtolength{\textheight}{-12cm}   









\bibliographystyle{ieeetr}
\bibliography{main}

\end{document}